\documentclass[runningheads]{llncs}
\usepackage{amsmath,amssymb}
\usepackage{graphicx}
\usepackage{multirow}
\usepackage{tikz}
\usepackage{comment} 
\usepackage{color}
\usepackage{xcolor}
\usepackage[numbers,sort&compress]{natbib}
\usepackage{booktabs}
\usepackage{url}
\usepackage{hyperref}
\usepackage[ruled,vlined]{algorithm2e}

\usepackage{soul}
\usepackage{amsmath}
\usepackage{amssymb}
\usepackage{microtype}
\usepackage{wrapfig}

\def\figurePath{figures/}

\def\myfigure#1#2{%
    \vspace{-.3cm}%
    \begin{figure}[htb]%
    \centering\includegraphics*[width = \linewidth]{\figurePath#1}%
    \vspace{-.4cm}%
    \caption{#2}%
    \vspace{-.3cm}%
    \label{fig:#1}%
    \end{figure}%
}

\newcommand{\mywrapfigure}[3]{%
\begin{wrapfigure}{r}{#2\columnwidth}%
\vspace{-.8cm}%
  \begin{center}%
    \includegraphics[width=#2\columnwidth]{\figurePath#1}%
    \vspace{-.3cm}%
    \caption{#3}%
    \label{fig:#1}%
    \vspace{-.8cm}%
  \end{center}%
\end{wrapfigure}%
\leavevmode%
}

\newcommand{\eg}{e.\,g., }
\newcommand{\ie}{i.\,e., }
\newcommand{\etal}{et~al.\ }

\newcommand{\argmin}[1]{\underset{#1}{\operatorname{arg\,min}\ }}

\newcommand{\refSec}[1]{Sec.~\ref{sec:#1}}
\newcommand{\refFig}[1]{Fig.~\ref{fig:#1}}
\newcommand{\refEq}[1]{Eq.~\ref{eq:#1}}
\newcommand{\refTbl}[1]{Tbl.~\ref{tbl:#1}}

\soulregister\ref7
\soulregister\cite7
\soulregister\refFig7
\soulregister\cite7
\soulregister\ref7
\soulregister\pageref7
\soulregister\shortcite7
\soulregister\eg0
\soulregister\ie0
\soulregister\etal0

\DeclareGraphicsExtensions{.png,.jpg,.pdf,.ai,.psd}
\newcommand{\mysection}[2]{\section{#1}\label{sec:#2}}
\newcommand{\mysubsection}[2]{\subsection{#1}\label{sec:#2}}
\newcommand{\mysubsubsection}[1]{\vspace{.2cm}\noindent{\textbf{#1}}\ }

\usepackage[width=122mm,left=12mm,paperwidth=146mm,height=193mm,top=12mm,paperheight=217mm]{geometry}

\begin{document}
\pagestyle{headings}
\mainmatter
\def\ECCVSubNumber{2919}

\renewcommand\AlCapFnt{\centering}
\renewcommand\AlCapNameFnt{\AlCapFnt}
\def\UrlBreaks{\do\/\do-}

\newcommand{\expected}{{\mathbb E}}
\newcommand{\numberOfCenters}{{n_\mathrm c}}
\newcommand{\numberOfScenes}{{n_\mathrm s}}
\newcommand{\scene}{\mathbf x}
\newcommand{\metric}{H}
\newcommand{\displacedScene}{\mathbf y}
\newcommand{\boxCenters}{\mathbf c}
\newcommand{\sceneNetwork}{{\mathcal S}}
\newcommand{\minisceneNetwork}{{\mathcal M}}
\newcommand{\sceneNetworkParameters}{\theta}
\newcommand{\lossNetwork}{{\mathcal L}}
\newcommand{\lossNetworkParameters}{\phi}
\newcommand{\oracle}{{\nabla}}
\newcommand{\boxCenter}{{\mathbf x}}
\newcommand{\cut}{{\mathcal C}}
\newcommand{\object}{{\mathbf o}}
\newcommand{\objectOffset}{\mathbf d}
\newcommand{\occupancyNetwork}{{\mathcal O}}
\renewcommand{\property}{\mathbf q}
\newcommand{\propertyNetwork}{{\mathcal P}}
\newcommand{\scenePropertyNetwork}{{\mathcal T}}
\newcommand{\occupancyThreshold}{\tau_\mathrm o}
\newcommand{\nonMaximaThreshold}{\tau_\mathrm m}
\newcommand{\result}{\mathbf r}

\newcommand\blfootnote[1]{%
  \begingroup
  \renewcommand\thefootnote{}\footnote{#1}%
  \addtocounter{footnote}{-1}%
  \endgroup
}

\definecolor{supervisedColor}{HTML}{4285f4}
\definecolor{slideColor}{HTML}{ea4335}
\definecolor{ourColor}{HTML}{fbbc04}
\definecolor{s3dColor}{HTML}{9900ff}
\definecolor{scanNetColor}{HTML}{93c47d}
\newcommand{\methodHead}[3]{
    \multicolumn{#1}{c}{
        \textsc{\textcolor{#2}{#3}}
    }
}
\newcommand{\missing}{\multicolumn{1}{c}{---$^1$}}
\newcommand{\classhead}[1]{\multicolumn{1}{c}{\rotatebox{85}{\texttt{\scriptsize{#1}}}}}

\title{Finding Your (3D) Center:\\3D Object Detection Using a Learned Loss}

\titlerunning{Finding Your (3D) Center}
%
\author{David Griffiths*\orcidID{0000-0002-8582-138X} \and
Jan Boehm\orcidID{0000-0003-2190-0449} \and
Tobias Ritschel}
\authorrunning{Griffiths D., Boehm J., Ritschel T.}
%
\institute{University College London, London, UK \\
\email{\{david.griffiths.16, j.boehm, t.ritschel\}@ucl.ac.uk}}

\pdfstringdefDisableCommands{%
  \def\\{}%
  \def\texttt#1{<#1>}%
}

\maketitle

\begin{abstract}
Massive semantically labeled datasets are readily available for 2D images, however, are much harder to achieve for 3D scenes.
Objects in 3D repositories like ShapeNet are labeled, but regrettably only in isolation, so without context.
3D scenes can be acquired by range scanners on city-level scale, but much fewer with semantic labels.
Addressing this disparity, we introduce a new optimization procedure, which allows training for 3D detection with raw 3D scans while using as little as 5\,\% of the object labels and still achieve comparable performance. 
Our optimization uses two networks.
A \emph{scene network} maps an entire 3D scene to a set of 3D object centers.
As we assume the scene not to be labeled by centers, no classic loss, such as Chamfer can be used to train it.
Instead, we use another network to emulate the loss.
This \emph{loss network} is trained on a small labeled subset and maps a non-centered 3D object in the presence of distractions to its own center.
This function is very similar -- and hence can be used instead of -- the gradient the supervised loss would provide.
Our evaluation documents competitive fidelity at a much lower level of supervision, respectively higher quality at comparable supervision. Supplementary material can be found at: \url{dgriffiths3.github.io}

\end{abstract}

\mysection{Introduction}{Introduction}
\blfootnote{* Corresponding author.}
We can reason about one 3D chair as we do about a 2D chair image, however, we cannot yet machine-understand a point cloud of a 3D room as we would do for a 2D room image.
For 2D images, massive amounts of manual human labeling have enabled amazing state-of-the-art object detectors \cite{redmon2016you,girshick2015fast,liu2016ssd,zhou2019objects}.
We also have massive repositories of clean 3D objects \cite{chang2015shapenet} which we can classify thanks to deep 3D point processing \cite{qi2017pointnetplusplus}.
But we do not have, despite commendable efforts \cite{song2015sun,dai2017scannet}, and probably might never have, 3D scene labeling at the extent of 2D images.
We hence argue that progress in 3D understanding even more critically depends on reducing the amount of supervision required.

While general unsupervised detection is an elusive goal, we suggest taking a shortcut: while we do not have labeled 3D scenes, we do have labeled 3D objects.
The key idea in this work is to first teach a \textit{loss network} everything that can be learned from seeing snippets of labeled objects. Next, we use this network to learn a \textit{scene network} that explores the relation of objects within scenes, but without any scene labels, \ie on raw scans.

After reviewing previous work, we will show how this cascade of networks is possible when choosing a slightly more primitive loss than the popular Chamfer loss and we propose two network architectures to implement it.
Results show how a state-of-the-art, simple, fast and feed-forward 3D detection network can achieve similar Chamfer distance and mAP@.25 scores to a supervised approach, but with only 5\,\% of the labels. 

\mysection{Previous Work}{PreviousWork}
2D object detection has been addressed by deep-learning based approaches like Fast R-CNN \cite{girshick2015fast}, YOLO \cite{redmon2016you}, SSD \cite{liu2016ssd} or the stacked hourglass architecture \cite{newell2016stacked} with great success.

In early work \citet{song2014sliding} have extended sliding window-detection to a 3D representation using voxels with templates of Hough features and SVM classifiers.
This approach was later extended to deep templates \cite{song2016deep}.
Both approaches use fully-supervised training on object locations given by bounding boxes.
We compare to such a sliding window approach using a point-based deep template.
\citet{hou20193d} complement a voxel-based approach with color 2D image information which more easily represents finer details.

\citet{karpathy2013object} detect objects by over-segmenting the scene and classifying segments as objects based on geometric properties such as compactness, smoothness, etc.
Similarly, \citet{chen20153d} minimize other features to 3D-detect objects in street scans.

While modern software libraries make voxels simple to work with, they are limited in the spatial extent of scenes they can process, and the detail of the scene they can represent.
\citet{qi2017pointnet} were first to suggest an approach to work on raw point clouds.
3D object detection in point clouds is investigated by
\citet{qi2019deep} and \citet{engelcke2017vote3deep} map the scene to votes, then those votes are clustered and each cluster becomes a proposal.
The vectors pointing from a seed to a vote are similar to the loss network gradients proposed in our method, but for VoteNet, this is part of the architecture during training and testing while for us these vectors are only part of the training.
Finally, VoteNet is trained fully supervised with object positions.
The idea of merging 2D images and 3D processing is applicable to point clouds as well, as shown by \citet{ku2018joint} and \citet{qi2020imvotenet}.

\citet{zhou2018voxelnet} question the usefulness of points for 3D detection and have suggested to re-sample point clouds to voxels again.
Also \citet{chen2019fast} show how combining point inputs, volume convolutions and point proposals can lead to good results.
For a survey on 3D detection, also discussing trade-offs of points and voxels, see the survey by \citet{griffiths2019review}.

Our architecture is inspired by Fast R-CNN \cite{girshick2015fast,shi2019pointrcnn}, which regresses object proposals in one branch, warps them into a canonical frame and classifies them in the other branch.
Recently, \citet{yang2019learning} have shown how direct mapping of a point cloud to bounding boxes is feasible.
\citet{feng2019relation} train a network with supervision that makes multiple proposals individually and later reasons about their relation.
Also, \citet{zhou2019objects} first work on center points for object representation alone and later regress the 2D bounding box and all other object features from image content around the points.
All these works tend to first extract proposals in a learned fashion and then reason about their properties or relations in a second, learned step.
We follow this design for the scene network, but drive its learning in an entirely different, unsupervised, way.
Finally, all of these works require only one feed-forward point cloud network pass, a strategy we will follow as well.

Unrelated to 2D or 3D detection, \citet{adler2017solving} have proposed to replace the gradient computation in an optimization problem by a neural network.
In computer vision, this idea has been used to drive light field \cite{flynn2019deepview} or appearance synthesis \cite{maximov2019deep}.
We take this a step further and use a network to emulate the gradients in a very particular optimization: the training of another network.

\mysection{Our Approach}{OurApproach}

We learn two networks: a \textit{scene network} and a \textit{loss network} (\refFig{Architecture}).
The first (\refFig{Architecture}, bottom) is deployed, while the second (\refFig{Architecture}, top) is only used in training.

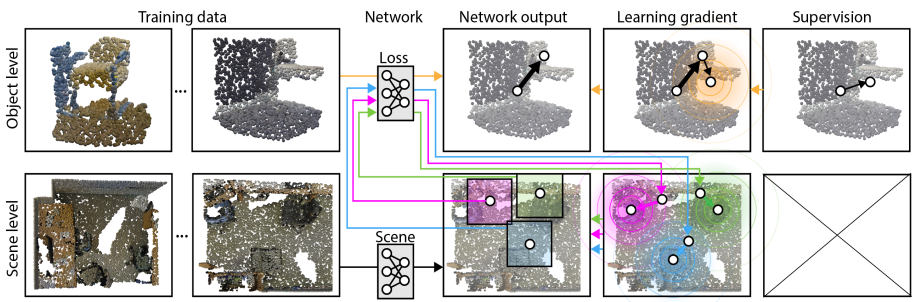
\myfigure{Architecture}{Our approach proceeds in two steps of training \textbf{(row)} with different training data (column one and two), networks (column three), outputs (column four), gradients (column five) and supervision (column six).
Object level training \textbf{(first row)} data comprises of 3D scene patches with known objects that are not centered.
The loss network maps off-center scenes to their center (big black arrow).
Its learning  follows the gradient of a quadratic potential (orange field) that has the minimum at the offset that would center the object.
This offset is the object-level supervision, as seen in the last column.
The scene network \textbf{(second row)} is trained to map a scene to all object centers, here for three chairs.
The gradient to train the scene network is computed by running the loss network from the previous step once for each object (here three times: blue, pink, green).
Note, that there is no scene-level supervision (cross).}

The \emph{scene network} maps 3D scenes to sets of 3D object centers.
The input data is a 3D point cloud.
The output is a fixed sized list of 3D object centers.
We assume a feed-forward approach, that does not consider any proposals \cite{hou20193d,newell2016stacked,song2014sliding,song2016deep} or voting \cite{qi2019deep,qi2020imvotenet}, but directly regresses centers from input data \cite{zhou2019objects,yang2019learning}.

The \emph{loss network} emulates the loss used to train the scene network. 
The input data is again a 3D point cloud, but this time of a single object, displaced by a random amount and subject to some other distortions.
Output is not the scalar loss, but the gradient of a Mean Squared Error loss function.

In the following, we will first describe the training (\refSec{Training}) before looking into the details of both the scene and loss network implementation (\refSec{Network}).

\mysubsection{Training}{Training}
The key contribution of our approach is a new way of training.
We will first look into a classic baseline with scene-level supervision, then introduce a hypothetical oracle that solves almost the same problem and finally show how this problem can be solved without scene-level supervision by our approach.

\myfigure{Overview}{
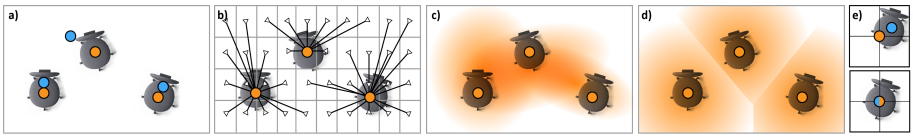
\textbf{a)} A 2D scene with three chair objects, supervised by centers (orange) and their predictions (blue).
\textbf{b)} The same scene, with the vector field of the oracle $\oracle$ shown as arrows.
\textbf{c)} A 2D Slice through a 6D cost function.
\textbf{d)} A 2D Slice through an alternative cost function, truncated at the Voronoi cell edges.
The oracle is the gradient of this.
\textbf{e)} The simple task of the loss network: given a chair not in the center (top), regress an offset such that it becomes centered.
}

\mysubsubsection{Supervised}
Consider learning the parameters $\sceneNetworkParameters$ of a scene network $\sceneNetwork_\sceneNetworkParameters$ which regresses object centers $\sceneNetwork_\sceneNetworkParameters(\scene_i)=\mathbf{\hat c}$ from a scene $\scene_i$.
The scene is labeled by a set of 3D object centers $\boxCenters_i$ (\refFig{Overview}, a).
This is achieved by minimizing the expectation
\begin{align}
\label{eq:Supervised}
\argmin{\sceneNetworkParameters}
\expected_{i}[
\metric(
\sceneNetwork_\sceneNetworkParameters(\scene_i)-
\boxCenters_{i}
)
]
,
\end{align}
using a two-sided Chamfer loss between the label point set $\mathbf c_i$ and a prediction $\mathbf{\hat c_i}$
\begin{align}
\metric(
\mathbf{\hat c},
\mathbf c
)=
\expected_i
[
\min_j
||
\mathbf{\hat c}_i
-
\mathbf c_j
||_2^2
]
+
\expected_i
[
\min_j
||
\mathbf c_i
-
\mathbf{\hat c}_j
||_2^2
].
\end{align}

\mywrapfigure{Chamfer}{0.25}{Chamfer loss.}

Under $\metric$, the network is free to report centers in any order, and ensures all network predictions are close to a supervised center (precision) and all supervised centers are close to at least one network prediction (recall) (\refFig{Chamfer}).

In this work, we assume the box center supervision $\boxCenters_i$ to not be accessible.
Tackling this, we will first introduce an oracle solving a similar problem. 

\mysubsubsection{Oracle}
Consider, instead of supervision, an \emph{oracle} function $\oracle(\scene)$ which returns for a 3D scene $\mathbf p$ \emph{the smallest offset by which we need to move the scene so that the world center falls onto an object center} (\refFig{Overview}, b).
Then, learning means to 
\begin{align}
\argmin{\theta}
\label{eq:Oracle}
\expected_{i, j}
[
||
\oracle(
\underbrace{
\scene_i\ominus
\sceneNetwork_\sceneNetworkParameters(\scene_i)_j}_{\displacedScene_{\sceneNetworkParameters, i, j}}
)
||_2^2
]
,
\end{align}
where $\scene\ominus\objectOffset$ denotes shifting a scene $\scene$ by an offset $\objectOffset$. 
The relation between \refEq{Supervised} and \refEq{Oracle} is intuitive: knowing the centers is very similar to pointing to the nearest center from every location.
It is, however, not quite the same.
It assures every network prediction would map to a center, but does not assure, that there is a prediction for every center.
We will need to deal with this concern later, by assuring space is well covered, so that there are enough predictions such that at least one maps to every center. 
We will denote a scene $i$ shifted to be centered around object $j$ by a \emph{scene network} with parameters $\sceneNetworkParameters$ as $\displacedScene_{\sceneNetworkParameters, i, j}$.

Every location that maps to itself, \ie a \emph{fixed point} \cite{FixedPoint} of $\oracle$, is an object center.
Hence, we try to get a scene network that returns the roots of the gradient field of the distance function around each object center (\refFig{Overview}, c):
\begin{align}
\argmin{\theta}
\expected_{i,j}
[
||
\oracle(
\displacedScene_{\sceneNetworkParameters, i, j}
)
||_2^2
]
.
\end{align}

\mysubsubsection{Learned loss}
The key idea is to emulate this oracle with a \emph{loss} network $\lossNetwork_\lossNetworkParameters$ having parameters $\lossNetworkParameters$ as in
\begin{align}
\label{eq:UnsupervisedSceneNetwork}
\argmin{\sceneNetworkParameters}
\expected_{i,j}
[
||
\lossNetwork_\lossNetworkParameters(
\displacedScene_{\sceneNetworkParameters, i, j}
)
||_2^2
]
.
\end{align}

The loss network does not need to understand any global scene structure, it only locally needs to \emph{center} the scene around the nearest object (\refFig{Overview}, d).
This task can be learned by working on local 3D object \emph{patches}, without scene-level supervision.
So we can train the loss network on any set of objects $\object_k$, translated by a known offset $\objectOffset_k$ using
\begin{align}
\label{eq:PatchLoss}
\argmin{\lossNetworkParameters}
\expected_k
[
||
\objectOffset_k-
\lossNetwork_\lossNetworkParameters(\object_k\ominus
\objectOffset_k)
||_2
]
.
\end{align}

As the loss network is local, it is also only ever trained on 3D patches.
These can be produced in several different ways: sampling of CAD models, CAD models with simulated noise, by pasting simulated results on random scene pieces, etc.
In our experiments, we use a small labeled scene subset to extract objects as follows: we pick a random object center and a 3D box of 1 meter size such that at least point representing an object surface is present in the box. 
Hence the center of the object is offset by a random, but known $\objectOffset_k$ we regress and subject to natural clutter.
Note, that the box does not, and does not have to, strictly cover the entire object -- which are of different sizes -- but has to be just large enough to guess the center. Alg. \ref{training_alg} demonstrates how the \emph{loss network} output can be used to provide \emph{scene network} supervision.

\begin{wrapfigure}{L}{0.46\textwidth}
\begin{minipage}{0.46\textwidth}
\begin{algorithm}[H]
\label{training_alg}
\SetAlgoLined
 $\mathcal L_\phi: \mathbb R^{n\times 3} \rightarrow \mathbb R^3$\;
 $\mathcal S_\theta: \mathbb R^{m\times 3} \rightarrow \mathbb R^{k\times 3}$\;
 $\mathtt{crop} : \mathbb R ^{m\times 3} \rightarrow \mathbb R^{n\times 3}$\;
 \While{loss training}{
  $x = \mathtt{sampleScene()}$\;
  $o = \mathtt{randObjectCenter()}$\;
  $d = \mathtt{randOffset()}$\;
  $p = \mathtt{crop}(x\ominus (o+d))$\;
  $\nabla = \frac \partial{\partial_\phi}||\mathcal L_\phi(p) - d||_2^2$\;
  $\phi = \mathtt{optimizer}(\phi, \nabla)$\;
 }
\
\While{scene training}{
  $x = \mathtt{sampleScene()}$\;
  $c = \mathcal S_\theta(x)$\;
  \For{$i=1\ldots k$}{
    $p = \mathtt{crop}(x\ominus c_i)$\;
    $\nabla_{i} = \mathcal L_\phi(p)$\;
  }
  $\theta = \mathtt{optimizer}(\theta, \nabla)$\;
 }
\caption{$\mathcal L$: loss network, $\mathcal S$: scene network, $k$: proposal count, $n$ 3D patch point count,  $m$ scene point count.}
\end{algorithm}
\end{minipage}
\end{wrapfigure}

\mysubsubsection{Varying object count}
The above was assuming the number of objects $\numberOfCenters$ to be known.
It did so when assuming a vector of a known dimension as supervision in \refEq{Supervised} and did so, when assuming the oracle \refEq{Oracle} and its derivations were returning gradient vectors of a fixed size.
In our setting this number is unknown.
We address this by bounding the number of objects and handling occupancy \ie a weight indicting if an object is present or not, at two levels.

First, we train an occupancy branch $\occupancyNetwork_\lossNetworkParameters$ of the loss network that classifies occupancy of a single patch, much like the loss network regresses the center.
We define space to be \emph{occupied}, if the 3D patch contains any points belonging to the given objects surface.
This branch is trained on the same patches as the loss network plus an equal number of additional 3D patches that do not contain any objects i.e. occupancy is zero.

Second, the occupancy branch is used to support the training of the scene network which has to deal with the fact that the number of actual centers is lower than the maximal number of centers.
This is achieved by ignoring the gradients to the scene networks parameters $\sceneNetworkParameters$ if the occupancy network reports the 3D patch about a center to not contain an object of interest.
So instead of \refEq{UnsupervisedSceneNetwork}, we learn
\begin{align}
\label{eq:MaskedUnsupervisedSceneNetwork}
\argmin{\sceneNetworkParameters}
\expected_{i,j}
[
\occupancyNetwork_\lossNetworkParameters(
\displacedScene_{\sceneNetworkParameters, i, j})
\lossNetwork_\lossNetworkParameters(
\displacedScene_{\sceneNetworkParameters, i, j})
].
\end{align}
The product in the sum is zero for centers of 3D patches that the loss network thinks, are not occupied and hence should not affect the learning.

\paragraph{Overlap}
When neither object centers nor their count are known, there is nothing to prevent two network outputs to map to the same center.
While such duplicates can to some level be addressed by non-maximum suppression as a (non-differentiable) post-process  to testing, we have found it essential to already prevent them (differentiable) from occurring  when training the scene network.
Without doing so, our training degenerates to a single proposal.

To this end, we avoid \emph{overlap}.
Let $v(q_1,q_2)$ be a function that is zero if the bounding boxes of the object in the scene centers do not overlap, one if they are identical and otherwise be the ratio of intersection.
We then optimize
\begin{align}
\argmin{\sceneNetworkParameters}
c_1(\sceneNetworkParameters)=
\expected_{i,j,k}
\left[
\occupancyNetwork_\lossNetworkParameters(
\displacedScene_{\sceneNetworkParameters, i, j})
\lossNetwork_\lossNetworkParameters(
\displacedScene_{\sceneNetworkParameters, i, j})
+
v(
\displacedScene_{\sceneNetworkParameters, i, j},
\displacedScene_{\sceneNetworkParameters, i, k})
\right].
\end{align}

We found that in case of a collision instead of mutually repelling all colliding objects, it can be more effective if out of multiple colliding objects, the collision acts on all but one winner object (winner-takes-all).
To decide the winner, we again use the gradient magnitude: if multiple objects collide, the one that is already closest to the target \ie the one with the smallest gradient, remains unaffected ($v=0$) and takes possession of the target, while all others adapt.

\mysubsubsection{Additional features}
For other object properties such as size, orientation, class of object, etc.\ we can proceed in two similar steps.
First, we know the object-level property vector $\property$, so we can train a \emph{property branch} denoted $\propertyNetwork_\sceneNetworkParameters$ that shares parameters $\sceneNetworkParameters$ with the loss network to regresses the property vector from the same displaced 3D patches as in \refEq{PatchLoss}
\begin{align}
\argmin{\lossNetworkParameters}
\expected_k
[
||
\property_k-
\propertyNetwork_\lossNetworkParameters(\object_k\ominus
\objectOffset_k)
||_1
]
.
\end{align}

For scene-level learning we extend the scene network by a branch $\scenePropertyNetwork_\sceneNetworkParameters$ to emulate what the property network had said about the 3D patch at each center, but now with global context and on a scene-level 
\begin{align}
\argmin{\sceneNetworkParameters}
c_1(\sceneNetworkParameters)
+
\alpha
\cdot
\expected_{i,j}
[
|
\scenePropertyNetwork_\sceneNetworkParameters(
\displacedScene_{\sceneNetworkParameters, i, j})
-
\propertyNetwork\lossNetworkParameters(
\displacedScene_{\sceneNetworkParameters, i, j})
|_1
]
.
\end{align}
For simplicity, we will denote occupancy just as any other object property and assume it to be produced by $\scenePropertyNetwork$ just, that it has a special meaning in training as defined in \refEq{MaskedUnsupervisedSceneNetwork}. 
We will next detail the architecture of all networks.

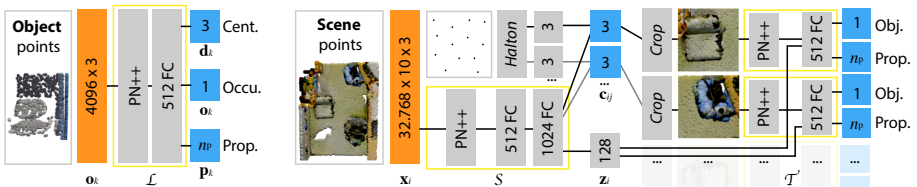
\myfigure{Details}{The object \textbf{(left)} and scene \textbf{(right)} network.
Input denoted orange, output blue, trainable parts yellow, hard-coded parts in italics.
Please see \refSec{Network} for a details.}%

\mysubsection{Network}{Network}
Both networks are implemented  using PointNet++ \cite{qi2017pointnetplusplus} optimized using ADAM.
We choose particularly simple designs and rather focus on the analysis of changes from the levels of supervision we enable.

\paragraph{Loss and occupancy network}
The loss network branches $\lossNetwork$ and $\occupancyNetwork$ share parameters $\lossNetworkParameters$ and both map 4,096 3D points to a 3D displacement vector, occupancy and other scalar features (left in \refFig{Details}).

\paragraph{Scene network}
The scene network branches $\sceneNetwork$ and $\scenePropertyNetwork$ jointly map a point cloud to a vector of 3D object centers and property vectors (including occupancy), sharing parameters $\sceneNetworkParameters$.
The box branch $\sceneNetwork$ first generates positions, next the scene is cropped around these positions and each 3D patch respectively fed into a small PointNet++ encoder $\minisceneNetwork$ to produce crop specific local feature encodings. Finally, we concatenate the global scene latent code $\sceneNetwork_z$ with the respective local latent code $\minisceneNetwork_z$ and pass it through the scene property branch $\scenePropertyNetwork$ MLP. 

The scene property branch is trained sharing all weights across all instances for all objects.
This is intuitive, as deciding that \eg a chair's orientation is the same for all chairs (the back rest is facing backwards), can at the same time be related to global scene properties (alignment towards a table).

Instead of learning the centers, we learn the residual relative to a uniform coverage of 3D space such that no object is missed during training.
The Hammersley pattern \cite{Hammersley} assures that, no part of 3D space is left uncovered.

We assume a fixed number of 32,768 input points for one scene.
Note, that we do not use color as input, a trivial extension.
Each MLP sub-layer is an MLP consisting of 3 fully-connected layers where layer 1 has 512 hidden states and the final layer contains the branch specific output nodes.

\paragraph{Post-process}
Our scene network returns a set of oriented bounding boxes with occupancy.
To reduce this soft answer to a set of detected objects, \eg to compute mAP metrics, we remove all bounding boxes with occupancy below a threshold $\occupancyThreshold$, which we set to 0.9 in all our results.

In the evaluation, the same will be done for our ablations \textsc{Sliding} and \textsc{Supervised}, just that these also require additional non-maximum suppression (NMS) as they frequently propose boxes that overlap.
To construct a final list of detections, we pick the proposal with maximal occupancy and remove any overlapping proposal with IoU $>.25$ and repeat until no proposals remain.

\mysection{Evaluation}{Evaluation}

We compare to different variants of our approach under different metrics and with different forms of supervision as well as to other methods.

\mysubsection{Protocol}{Protocol}

\mysubsubsection{Data sets}

We consider two large-scale sets of 3D scanned scenes: Stanford 2D-3D-S dataset (S3D) \cite{armeni2017joint} and ScanNet \cite{dai2017scannet}.
From both we extract, for each scene, the list of object centers and object features for all objects of one class.

We split the dataset in three parts (\refFig{LabelRatio}):
First, the test dataset is the official test dataset (pink in \refFig{LabelRatio}).
The remaining training data is split into two parts: a labeled, and and unlabeled part.
The labeled part (orange in \refFig{LabelRatio}) has all 3D scenes with complete annotations on them.
The unlabeled part (blue in \refFig{LabelRatio}) contains only raw 3D point cloud without annotation. 
Note, that the labeled data is a subset of the unlabeled data, not a different set.
\mywrapfigure{LabelRatio}{0.35}{Label ratio.}%
We call the ratio of labeled over unlabeled data the \emph{label ratio}.
To more strictly evaluate transfer across data sets, we consider ScanNet completely unlabeled.
All single-class results are reported for the class \texttt{chair}.

\mysubsubsection{Metrics}
Effectiveness is measured using the Chamfer distance (less is better) also used as a loss in \refEq{Supervised} and the established mean Average Precision mAP@.25, (more is better) of a x\,\% bounding box overlap test.
X is chosen at 25\,\%.

\mysubsubsection{Methods}
We consider the following \emph{three} methods:
\textsc{Supervised} is the supervised approach define by \refEq{Supervised}.
This method can be trained only on the labeled part of the training set.
\textsc{Sliding} window is an approach that applies our loss network, trained on the labeled data, to a dense regular grid of 3D location in every 3D scene to produce a heat map from which final results are generated by NMS.
\textsc{Ours} is our method.
The loss network is trained on the labeled data (orange in \refFig{LabelRatio}).
The scene network is trained on the unlabeled data (blue in \refFig{LabelRatio}), which includes the labeled data (but without accessing the labels) as a subset.

\mysubsection{Results}{Results}

\mysubsubsection{Effect of supervision}
The main effect to study is the change of 3D detection quality in respect to the level of supervision.
In \refTbl{Main}, different rows show different label ratios.
The columns show Chamfer error and mAP@.25 for the class \texttt{chair} trained and tested on S3D.

\begin{table}
    \vspace{-.5cm}%
    \centering%
    \caption{Chamfer error (less is better) and mAP@.25 (more is better) (\textbf{columns}), as a function of supervision (\textbf{rows}) in units of label ratio on the S3D class \texttt{chair}.
    Right, the supervision-quality-relation plotted as a graph for every method (color).}%
    \label{tbl:Main}%
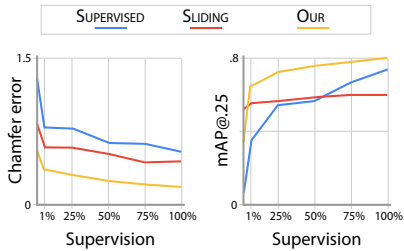
    \begin{minipage}{0.55\linewidth}
        \setlength{\tabcolsep}{4.0pt}%
        \begin{tabular}{rrrrrrrr}
        &
        \multicolumn{3}{c}{Chamfer error}&
        \multicolumn{3}{c}{mAP}
        \\
        \cmidrule(lr){2-4}
        \cmidrule(lr){5-7}
        Ratio&
        \methodHead{1}{supervisedColor}{Sup}&
        \methodHead{1}{slideColor}{Sli}&
        \methodHead{1}{ourColor}{Our}&
        \methodHead{1}{supervisedColor}{Sup}&
        \methodHead{1}{slideColor}{Sli}&
        \methodHead{1}{ourColor}{Our}
        \\
        \toprule
1\,\% & 1.265 & .850 & \textbf{.554} & .159 & \textbf{.473} & .366 \\
5\,\% & .789 & .577 & \textbf{.346} & .352 & .562 & \textbf{.642} \\
25\,\% & .772 & .579 & \textbf{.274} & .568 & .573 & \textbf{.735} \\
50\,\% & .644 & .538 & \textbf{.232} & .577 & .589 & \textbf{.773} \\
75\,\% & .616 & .437 & \textbf{.203} & .656 & .592 & \textbf{.785} \\
100\,\% & .557 & .434 & \textbf{.178} & .756 & .598 & \textbf{.803}\\
        \bottomrule
        \end{tabular}%
    \end{minipage}%
    \begin{minipage}{0.45\linewidth}
        \includegraphics[width=1.0\linewidth]{figures/MainTable}%
    \end{minipage}%
    \vspace{-.5cm}%
\end{table}%

We notice that across all levels of supervision, \textsc{Our} approach performs better in Chamfer error and mAP than \textsc{Sliding} window using the same object training or \textsc{Supervised} training of the same network.
It can further be seen, how all methods improve with more labels.
Looking at a condition with only 5\,\% supervision, \textsc{Our} method can perform similar to a \textsc{Supervised} method that had 20$\times$ the labeling effort invested.
At this condition, our detection is an acceptable $.642$, which \textsc{Supervised} will only beat when at least 75\,\% of the dataset is labeled.
It could be conjectured, that the scene network does no more than emulating to slide a neural-network object detector across the scene.
If this was true, \textsc{Sliding} would be expected to perform similar or better than \textsc{Ours}, which is not the case.
This indicates, that the scene network has indeed learned something not known at object level, something about the relation of the global scene without ever having labels on this level.

\mywrapfigure{Histograms}{0.4}{Error distribution.}
\refFig{Histograms} plots the rank distribution (horizontal axis) of Chamfer distances (vertical axis) for different methods (colors) at different levels of supervision (lightness).
We see that \textsc{Our} method performs well across the board, \textsc{Supervised} has a more steep distribution compared to \textsc{Sliding}, indicating it produces good as well as bad results, while the former is more uniform.
In terms of supervision scalability, additional labeling invested into our method (brighter shades of yellow) result in more improvements to the right side of the curve, indicating, additional supervision is reducing high error-responses while already-good answers remain.

\mysubsubsection{Transfer across data sets}
So far, we have only considered training and testing on S3D.
In \refTbl{Transfer}, we look into how much supervision scaling would transfer to another data set, ScanNet.
Remember, that we treat ScanNet as unlabeled, and hence, the loss network will be strictly only trained on objects from S3D.
The three first rows in \refTbl{Transfer} define the conditions compared here: a loss network always trained on S3D, a scene network trained on either S3D or ScanNet and testing all combinations on both data sets.

\begin{table}
    \vspace{-.5cm}
    \centering%
    \caption{Transfer across data sets:
    Different rows show different levels of supervision, different columns indicate different methods and metrics.
    The plot on the right visualizes all methods in all conditions quantified by two metrics.
    Training either on S3D or on ScanNet.
    The metrics again are Chamfer error (also the loss) and mAP@.25.
    Colors in the plot correspond to different training, dotted/solid to different test data.}%
    \label{tbl:Transfer}%
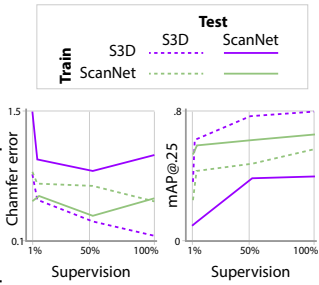
    \begin{minipage}{0.65\linewidth}
        \setlength{\tabcolsep}{1.75pt}%
        \begin{tabular}{r rr rr rr rr}
        \textit{Loss:}&
        \multicolumn{4}{c}{S3D}&
        \multicolumn{4}{c}{S3D}
        \\
        \textit{Scene:}&
        \methodHead{4}{s3dColor}{S3D}&
        \methodHead{4}{scanNetColor}{ScanNet}
        \\
        \cmidrule(lr){2-5}        
        \cmidrule(lr){6-9}        
        \textit{Test:}&
        \multicolumn{2}{c}{S3D}&
        \multicolumn{2}{c}{ScanNet}&
        \multicolumn{2}{c}{S3D}&
        \multicolumn{2}{c}{ScanNet}
        \\
        \cmidrule(lr){2-3}        
        \cmidrule(lr){4-5}        
        \cmidrule(lr){6-7}        
        \cmidrule(lr){8-9}        
        Ratio&
        Err.&
        mAP&
        Err.&
        mAP&
        Err.&
        mAP&
        Err.&
        mAP
        \\
        \toprule
1\% & 0.554 & .366 & 1.753 & .112 & 0.579 & .296 & 0.337 & .548 \\
5\% & 0.346 & .642 & 0.727 & .138 & 0.466 & .463 & 0.703 & .599 \\
50\% & 0.232 & .773 & 0.588 & .380 & 0.447 & .497 & 0.258 & .645 \\
100\% & 0.178 & .803 & 0.789 & .384 & 0.336 & .555 & 0.356 & .661 \\
        \bottomrule
        \end{tabular}%
    \end{minipage}%
    \begin{minipage}{0.35\linewidth}
        \includegraphics[width=1.0\linewidth]{figures/Transfer}%
    \end{minipage}%
    \vspace{-0.5cm}%
\end{table}%

Column two and three in \refTbl{Transfer} and the dotted violet line in the plot, iterate the scaling of available label data we already see in \refTbl{Main} when training and testing on S3D.
Columns four and five, show a method trained on S3D but tested on ScanNet.
We find performance to be reduced, probably, as the domain of ScanNet is different from the one of S3D.
If we include the unlabeled scenes of ScanNet in the training, as seen in columns six to nine, the quality increases again, to competitive levels, using only S3D labels and 0\,\% of the labels available for ScanNet.

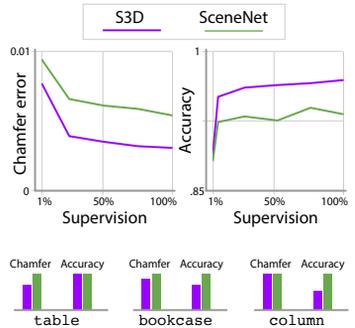
\begin{table}
    \centering%
    \caption{Performance of the loss network for different label ratio \textbf{(rows)} on different test data and according to different metrics \textbf{(columns)}. $^1$\textit{Class not present in ScanNet.}}%
    \label{tbl:LossNetwork}%
    \begin{minipage}{0.625\linewidth}
        \setlength{\tabcolsep}{1.75pt}%
        \begin{tabular}{r rr rr rr rr}
        &
        &
        &
        &
        \methodHead{2}{s3dColor}{S3D}&
        \methodHead{2}{scanNetColor}{ScanNet}
        \\
        \cmidrule(lr){5-6}        
        \cmidrule(lr){7-8}        
        \multicolumn{1}{c}{Ratio}&
        \multicolumn{1}{c}{Class}&
        \multicolumn{1}{c}{\#Sce}&
        \multicolumn{1}{c}{\#Obj}&
        \multicolumn{1}{c}{Err.}&
        \multicolumn{1}{c}{Acc.}&
        \multicolumn{1}{c}{Err.}&
        \multicolumn{1}{c}{Acc.}
       \\
        \toprule
1\,\% & \texttt{chair} & 11 & 2400 & .0085 & .853 & .0099 & .843 \\
5\,\% & \texttt{chair} & 54 & 16,000 & .0052 & .936 & .0075 & .899 \\
25\,\% & \texttt{chair} & 271 & 47,200 & .0049 & .949 & .0071 & .907 \\
50\,\% & \texttt{chair} & 542 & 121,191 & .0046 & .953 & .0069 & .902 \\
75\,\% & \texttt{chair} & 813 & 162,000 & .0045 & .955 & .0065 & .920 \\
100\,\% & \texttt{chair} & 1084 & 223,980 & .0043 & .960 & .0068 & .911 \\
\midrule
5\,\% & \texttt{table} & 54 & 5060 & .0078 & .921 & \missing & \missing \\
5\,\% & \texttt{bcase} & 54 & 4780 & .0093 & .819 & \missing & \missing \\
5\,\% & \texttt{column} & 54 & 2780 & .0100 & .855 & \missing & \missing \\
        \bottomrule
        \end{tabular}%
    \end{minipage}%
    \begin{minipage}{0.375\linewidth}
        \includegraphics[width=1.0\linewidth]{figures/LossNetworkTable}%
    \end{minipage}%
    \vspace{-.5cm}%
\end{table}%

\refTbl{LossNetwork} further illustrate the loss network: how good are we at finding vectors that point to an object center?
We see that the gradient error and the confidence error, both go down moderately with more labels when training and testing on S3D (violet).
The fact that not much is decreasing in the loss network, while the scene network keeps improving, indicates the object task can be learned from little data, and less object-level supervision is required than what can be learned on a scene level, still.
We further see, that the loss network generalizes between data sets from the fact that it is trained on S3D (violet curve) but when tested on ScanNet (green curve) goes down, too.

\begin{wraptable}{r}{5.5cm}%
    \vspace{-1cm}%
    \centering%
    \caption{Chamfer error and mAP@.25 reported for varying the number of scenes.}%
    \label{tbl:SceneScaling}%
    \setlength{\tabcolsep}{3.0pt}%
    \begin{minipage}{3cm}
    \begin{tabular}{r rr}
    &
    \methodHead{2}{ourColor}{Our}
    \\
    \cmidrule(lr){2-3}
    \#Sce&
    \multicolumn{1}{c}{Err.}&
    \multicolumn{1}{c}{mAP}
    \\
    \toprule
    66 & .643 & .079 \\
    330 & .509 & .242 \\
    1648 & .506 & .360 \\
    3295 & .457 & .412 \\
    4943 & .435 & .479 \\
    6590 & .407 & .599 \\
    \bottomrule
    \end{tabular}%
    \end{minipage}%
    \begin{minipage}{2.5cm}
        \includegraphics[width=2.5cm]{figures/SceneScaling}%
 \end{minipage}%
 \vspace{-.5cm}%
\end{wraptable}%

Besides seeing how quality scales with the amount of labeled supervision for training the loss network, it is also relevant to ask what happens when the amount of unlabeled training data for the scene network is increased while holding the labeled data fixed.
This is analyzed in \refTbl{SceneScaling}.
Here we took our loss network and trained it at 5\,\% label ratio on S3D and tested on ScanNet.
Next, the scene network was trained, but on various number of scenes from ScanNet, which, as we said, is considered unlabeled.
The number of scenes changes over columns, resp.\ along the horizontal axis in the plot.
We see that without investing any additional labeling effort, the scene network keeps increasing substantially, indicting what was learned on a few labeled S3D objects can enable understanding the structure of ScanNet.

\mysubsubsection{Different classes}
\refTbl{Main} was analyzing the main axis of contribution: different levels of supervision but for a single class. This has shown that at around a label ratio of 5\,\% \textsc{Our} method performs similar to a \textsc{Supervised} one.
Holding the label ration of 5\,\% fixed and repeating the experiment for other classes, is summarized in \refTbl{Classes}.
We see, that the relation between \textsc{Supervised}, \textsc{Sliding} and \textsc{Ours} is retained across classes.

\begin{table}[h]%
    \centering%
    \caption{Chamfer error (less is better) and mAP@.25 precision (more is better) \textbf{(columns)}, per class \textbf{(rows)} at a supervision of 5\,\% labeling ratio.}%
    \label{tbl:Classes}%
    \setlength{\tabcolsep}{3.0pt}%
    \begin{minipage}{0.55\linewidth}
    \begin{tabular}{rrrrrrrr}
    &
    \multicolumn{3}{c}{Chamfer error}&
    \multicolumn{3}{c}{mAP}
    \\
    \cmidrule(lr){2-4}
    \cmidrule(lr){5-7}
    Class&
    \methodHead{1}{supervisedColor}{Sup}&
    \methodHead{1}{slideColor}{Sli}&
    \methodHead{1}{ourColor}{Our}&
    \methodHead{1}{supervisedColor}{Sup}&
    \methodHead{1}{slideColor}{Sli}&
    \methodHead{1}{ourColor}{Our}
    \\
    \toprule
\texttt{chair} & 0.789 & 0.577 & .346 & .352 & .562 & .642 \\
\texttt{table} & 1.144 & 1.304 & .740 & .282 & .528 & .615 \\
\texttt{bookcase} & 1.121 & 1.427 & .979 & .370 & .298 & .640 \\
\texttt{column} & 0.900 & 2.640 & .838 & .490 & .353 & .654 \\
    \bottomrule
    \end{tabular}%
    \end{minipage}%
    \begin{minipage}{0.45\linewidth}
        \includegraphics[width=1.0\linewidth]{figures/ClassTable}%
 \end{minipage}%
\end{table}%

\mysubsubsection{Comparison to other work}
In \refTbl{Competitors} we compare our approach to other methods.
Here, we use 20\,\% of ScanNet V2 for testing and the rest for training.
Out of the training data, we train our approach once with 100\,\% labeled and once with only 5\,\% labeled.
Other methods were trained at 100\,\% label ratio.

\begin{table}
    \centering%
    \caption{Performance (mAP(\%) with IoU threshold .25) of different methods \textbf{(rows)} on all classes \textbf{(columns)} of ScanNet V2.
    $^1$\textit{5 images}.
    $^2$\textit{Only xyz}.
    $^3$\textit{Their ablation; similar to our backbone}.
    }%
    \tiny%
    \label{tbl:Competitors}%
    \setlength{\tabcolsep}{1.3pt}%
    \vspace{-.5cm}%
    \begin{tabular}{lc rrrrrrrrrrrrrrrrrrr}
        \\
    \scriptsize{Method}&&
    \classhead{cabinet} & 
    \classhead{bed} & 
    \classhead{chair} & 
    \classhead{sofa} & 
    \classhead{table} & 
    \classhead{door} & 
    \classhead{window} & 
    \classhead{bookshelf} & 
    \classhead{picture} & 
    \classhead{counter} & 
    \classhead{desk} & 
    \classhead{curtain} & 
    \classhead{fridge} & 
    \classhead{curtain} & 
    \classhead{toilet} & 
    \classhead{sink} & 
    \classhead{bathtub} & 
    \classhead{other}&
    \multicolumn{1}{c}{mAP}
    \\
    \toprule
3DSIS$^1$&\cite{hou20193d}& 19.8 & 69.7 & 66.2 & 71.8 & 36.1 & 30.6 & 10.9 & 27.3 & 0.0 & 10.0 & 46.9 & 14.1 & 53.8 & 36.0 & 87.6 & 43.0 & 84.3 & 16.2 & 40.2 \\
3DSIS$^2$&\cite{hou20193d}& 12.8 & 63.1 & 66.0 & 46.3 & 26.9 & 8.0 & 2.8 & 2.3 & 0.0 & 6.9 & 33.3 & 2.5 & 10.4 & 12.2 & 74.5 & 22.9 & 58.7 & 7.1 & 25.4 \\
MTML&\cite{lahoud20193d}& 32.7 & 80.7 & 64.7 & 68.8 & 57.1 & 41.8 & 39.6 & 58.8 & 18.2 & 0.4 & 18.0 & 81.5 & 44.5 & 100.0 & 100.0 & 44.2 & 100.0 & 36.4 & 54.9 \\
VoteNet&\cite{qi2019deep}& 36.3 & 87.9 & 88.7 & 89.6 & 58.8 & 47.3 & 38.1 & 44.6 & 7.8 & 56.1 & 71.7 & 47.2 & 45.4 & 57.1 & 94.9 & 54.7 & 92.1 & 37.2 & 58.7 \\
BoxNet$^3$&\cite{lahoud20193d}& \multicolumn{18}{c}{No per-class information available} & 45.4 \\
3D-BoNet&\cite{yang2019learning}& 58.7 & 88.7 & 64.3 & 80.7 & 66.1 & 52.2 & 61.2 & 83.6 & 24.3 & 55.0 & 72.4 & 62.0 & 51.2 & 100.0 & 90.9 & 75.1 & 100.0 & 50.1 & 68.7 \\
\midrule
Ours 100\,\%& & 43.0 & 70.8 & 58.3 & 16.0 & 44.6 & 28.0 & 13.4 & 58.2 & 4.9 & 69.9 & 74.0 & 75.0 & 36.0 & 58.9 & 79.0 & 47.0 & 77.9 & 48.2 & 50.2 \\
Ours \hphantom{00}5\,\% & & 38.1 & 68.9 & 58.9 & 88.8 & 42.5 & 21.1 & 9.0 & 53.2 & 6.8 & 53.9 & 68.0 & 62.3 & 26.5 & 45.6 & 69.9 & 40.4 & 66.9 & 48.0 & 48.3 \\
\bottomrule
    \end{tabular}%
    \vspace{-.5cm}%
\end{table}%

We see that our approach provides competitive performance, both at 100\,\% of the labels, as well as there is only a small drop when reducing supervision by factor 20$\times$.
Our mAP at 100\,\%  of the labels is better than both variants (with and without color) of 3DSIS \cite{hou20193d} from 2018 and similar to MTML \cite{lahoud20193d} from 2019.
VoteNet \cite{qi2019deep} and 3D-BoNet \cite{yang2019learning} are highly specialized architectures from 2019 that have a higher mAP.
We have included BoxNet from \citet{qi2019deep}, an ablation they include as a vanilla 3D detection approach that is similar to what we work with.
We achieve similar even slightly better performance, yet at 5\,\% of the supervision.
In some categories, our approach wins over all approaches.
We conclude that a simple backbone architecture we use is no contribution and cannot win over specialized ones, but that it also is competitive to the state-of-the-art. We should note here, as we do not carry out Semantic instance segmentation in our network, we did not test on the official test ScanNet benchmark test set. Instead, we reserve 20\% of the labeled training scenes for testing.

\mysubsubsection{Qualitative results}
\refFig{Qualitative} shows qualitative example results of our approach.
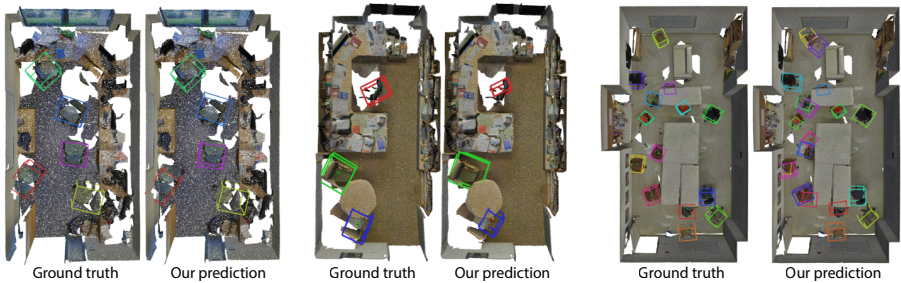
\myfigure{Qualitative}{Qualitative results of our approach and the ground truth for \texttt{chair} on S3D.}

\mysubsubsection{Computational efficiency}
Despite the additional complexity in training, at deployment, out network is a direct and fast forward architecture, mapping a point cloud to bounding boxes.
Finding 20 proposals in 32,768 points takes 189\,ms, while the supervised takes the same amount of time, with the small overhead of a NMS (190\,ms) on a Nvidia RTX 2080Ti.
Our CPU implementation of sliding window requires 4.7\,s for the same task on a i7-6850K CPU @ 3.60GHz.
All results are computed with those settings. 

\mysection{Discussion}{Discussion}

\paragraph{How can \textsc{Ours} be better than the \textsc{Supervised}?}
It is not obvious why at 100\,\% label ratio in \refTbl{Main}, the \textsc{Supervised} architecture performs at an mAP of .756 while \textsc{Ours} remains slightly higher at an mAP of .803.
This is not just variance of the mAP estimation (computed across many identical objects and scenes).

A possible explanation for this difference is, that our training is no drop-in replacement for supervised training.
Instead, it optimizes a different loss (truncation to the nearest object and collision avoidance) that might turn out to be better suited for 3D detection than what it was emulating in the beginning.
We, for example, do not require NMS.
As our training does not converge without those changes to the architecture, some of the effects observed might be due to differences in architecture and not due to the training.
We conjecture future work might consider exploring different losses, involving truncation and collision, even when labels are present.

\paragraph{Why Hammersley?}
Other work has reasoned about what intermediate points to use when processing point clouds.
When voting \cite{qi2019deep}, the argument is, that the centers of bounding boxes are not part of the point set, and hence using a point set that is any subset of the input is not a good solution.
While we do not vote, we also have chosen not to use points of the scene as the initial points.
We also refrain from using any improved sampling of the surface, such as Poisson disk \cite{hermosilla2018monte} sampling as we do not seek to cover any particular instance but space in general, covered by scenes uniformly. 

\paragraph{How can the scene network be ``better'' than the loss network?}
As the loss network is only an approximation to the true loss, one might ask, how a scene network, trained with this loss network, can perform better than the loss network alone, \eg how can it, consistently (\refTbl{Main}, \ref{tbl:Transfer}, \ref{tbl:SceneScaling} and \ref{tbl:Classes}), outperform \textsc{SlidingWindow}?

Let us assume, that a scene network trained by a clean supervision signal can use global scene structure to solve the task.
If now the supervision signal would start to be corrupted by noise, recent work has shown for images \cite{lehtinen2018noise2noise} or point clouds \cite{hermosilla2019total}, that a neural network trained under noise will converge to a result that is very similar to the clean result: under $\mathcal L_2$ it will converge to the mean of the noise, under $\mathcal L_1$ to its median, etc.
The amount of variance of that noise does not influence the result, what matters is that the noise is unbiased.
In our case, this means if we were to have supervision by noisy bounding boxes, that would not change anything, except that the scene network training would converge slower but still to the mean or median of that noise distribution, which is, the correct result.
So what was done in our training, by using a network to approximate the loss, means to just introduce another form of noise into the training.

\mysection{Conclusion}{Conclusion}
We have suggested  a novel training procedure to reduce the 3D labeling effort required to solve a 3D detection task.
The key is to first learn a loss function on a small labeled local view of the data (objects), which is then used to drive a second learning procedure to capture global relations (scenes).
The way to enlightenment here is to ``find your center'': the simple task of taking any piece of 3D~scene and shifting it so it becomes centered around the closest object.
Our analysis indicates that the scene network actually understands global scene structure not accessible to a sliding window.
Our network achieves state of the art results, executes in a fraction of a second on large point clouds with typically only 5\,\% of the labeling effort.
We have deduced what it means exactly to learn the loss function, the new challenges associated with this problem and proposed several solutions to overcome these challenges.

In future work, other tasks might benefit from similar decoupling of supervision labels and a learned loss, probably across other domains or modalities.

\clearpage

\bibliographystyle{splncsnatbib}
\bibliography{references}

\clearpage
\section*{Supplementary}\label{appendix}

We will here give further details of our architecture. For simplicity we ignore batch dimensions.
All branches for both \texttt{loss} and \texttt{scene} networks have a shared PointNet++ encoder shown in \refTbl{pnetencoder} with input 4,096 and 32,768 respectively.
We use the original ``Single Scale Grouping" architecture from PointNet++.
Unlike the original implementation no batch normalization, weight regularization or dropout layers are used.
We define the implemented architectures in \refTbl{pnetencoder} and \refTbl{mininetencoder}.

\begin{table}[h]
    \centering%
    \caption{PointNet++ encoder consisting of four ``Set Abstraction'' (SA) layers.
    Input is either 32,768 or 4,096 points for training \texttt{scene} or \texttt{loss} networks respectively.
    \textit{Points} indicates the number of points selected from the fartest point sampling algorithm.
    \textit{Samples} is the maximum points per neighborhood defined using a ball point query with radius $r$.}%
    \label{tbl:pnetencoder}%
        \begin{tabular}{rrrrrrrcrcl}
        \\
        Layer & 
        \multicolumn{3}{c}{MLP} & 
        Points & 
        Samples & 
        Radius & 
        Activation & 
        \multicolumn{3}{c}{Output Shape}\\ 
        \toprule
        Input & & & & & & & & $(32768 / 4096)$& $\times$& $3$ \\
        SA & [64,& 64,& 128] & 1024 & 32 & 0.1 & ReLU & $1024$& $\times$& $128$ \\
        SA & [64,& 64,& 128] & 256 & 32 & 0.2 & ReLU & $256$& $\times$& $128$ \\
        SA & [128,&128,& 256] & 64 & 64 & 0.4 & ReLU & $64$& $\times$& $256$ \\
        SA & [256,&512,& 128] & None & None & None & ReLU &  $1$& $\times$& $128$ \\
        \bottomrule
        \end{tabular}
\end{table}

Our \textit{local} encoder is a 3 layer SSG network, which typically consists of half the number of units for each respective MLP layer.

\begin{table}
    \caption{PointNet++ local encoder consisting of 3 SA layers. See \refTbl{pnetencoder} caption for further details regarding columns.}%
    \label{tbl:mininetencoder}%
    \centering%
        \begin{tabular}{rrrrrrrcrcl}
        \\
        Layer & 
        \multicolumn{3}{c}{MLP} & 
        Points & 
        Samples & 
        Radius & 
        Activation & 
        \multicolumn{3}{c}{Output Shape}
        \\ 
        \toprule
        Input & & & & & & & & $4096$& $\times$& $3$ \\
        SA & [32,& 32,& 64] & 1024 & 32 & 0.1 & ReLU & $256$& $\times$& $128$ \\
        SA & [64,& 64,& 128] & 256 & 32 & 0.2 & ReLU & $64$& $\times$& $256$ \\
        SA & [128,& 128,& 128] & None & None & None & ReLU & $1$& $\times$& $128$ \\
        \bottomrule
        \end{tabular}%
\end{table}%

Our loss network architecture has a shared PointNet++ encoder with three branches; gradient, property and objectness.
Each branch consists of two fully-connected layers with a ReLU activation for the hidden layer.
We use \texttt{softmax} for both single and multi-class examples.
For single class where $k=2$ the first class represents \texttt{empty} and the second class represents \texttt{present}.
In a multi-class setting $k$ is the number of classes.
We do not have a \texttt{empty} class for multi-class.
Both gradient and property branches do not have a final layer activation function.

\begin{table}
    \caption{Loss network architecture. PNet++ is defined in \refTbl{pnetencoder}.}%
    \label{tbl:lossnetarch}%
    \centering%
        \begin{tabular}{lccrrcl}
        \\
        Branch &
        Layer &
        Kernel &
        Activation &
        \multicolumn{3}{c}{Output Shape}
        \\
        \toprule
        Trunk & PNet++ & & ReLU & $1$& $\times$& $128$ \\
        \midrule
        
        \multirow{2}{*}{Gradient} 
        & {FC} & {$1\times1$} & {ReLU} & $1$& $\times$& $512$ \\
        & {FC} & {$1\times1$} & {None} & $1$& $\times$& $3$ \\
        \midrule
        \multirow{2}{*}{Property} 
        & {FC} & {$1\times1$} & {ReLU} & $1$& $\times$& $512$ \\
        & {FC} & {$1\times1$} & {None} & $1$& $\times$& $5$ \\

        \midrule
        \multirow{2}{*}{Objectness} 
        & {FC} & {$1\times1$} & {ReLU} & $1$& $\times$& $512$ \\
        & {FC} & {$1\times1$} & {Softmax} & $1$& $\times$& $k$ \\
    
        \bottomrule
        \end{tabular}

\end{table}%

The \textit{scene} network architecture is similar to the loss network but is designed for multiple proposals $n$ as apposed to the loss network where $n=1$. The property and objectness branches take patch codes generated by the local encoder. The patches are local crops from the center branch proposals. Note that the output of the PointNet++ encoder and local encoder are both $1 \times 128$. These latent codes are concatenated for input into the scene network property and objectness branches.

\begin{table}
    \caption{Scene network architecture. PNet++ and PNet++ local are defined in \refTbl{pnetencoder} and \ref{tbl:mininetencoder} respectively.}
    \label{tbl:scenenet_arch}
    \centering%
        \begin{tabular}{lcccrcl}
        \\
        Branch & 
        Layer & 
        Kernel & 
        Activation & 
        \multicolumn{3}{c}{Output Shape}
        \\ 
        \toprule
        Trunk & PNet++ encoder & & ReLU & $1$& $\times$& $128$ \\
        \toprule
        \multirow{2}{*}{Center} 
            & {FC} & {$1\times1$} & {ReLU} & $n$& $\times$& $512$ \\
             & {FC} & {$1\times1$} & {None} & $n$& $\times$& $3$ \\
        \toprule

        \multirow{3}{*}{Property} 
            & {PNet++ local} & {-} & {ReLU} & $n$& $\times$& $128$ \\
            & {FC} & {$1\times1$} & {ReLU} & $n$& $\times$& $512$ \\
            & {FC} & {$1\times1$} & {None} & $n$& $\times$& $5$ \\

        \toprule
    
        \multirow{3}{*}{Objectness}
        & {PNet++ local} & {-} & {ReLU} & $n$& $\times$& $128$ \\
        & {FC} & {$1\times1$} & {ReLU} & $n$& $\times$& $512$ \\
        & {FC} & {$1\times1$} & {Softmax} & $n$& $\times$& $k$ \\
    
        \bottomrule
        \end{tabular}

\end{table}%

All network weights were initialized using the Glorot Normal initialization. We do not employ regularization or batch normalization in any of the networks.

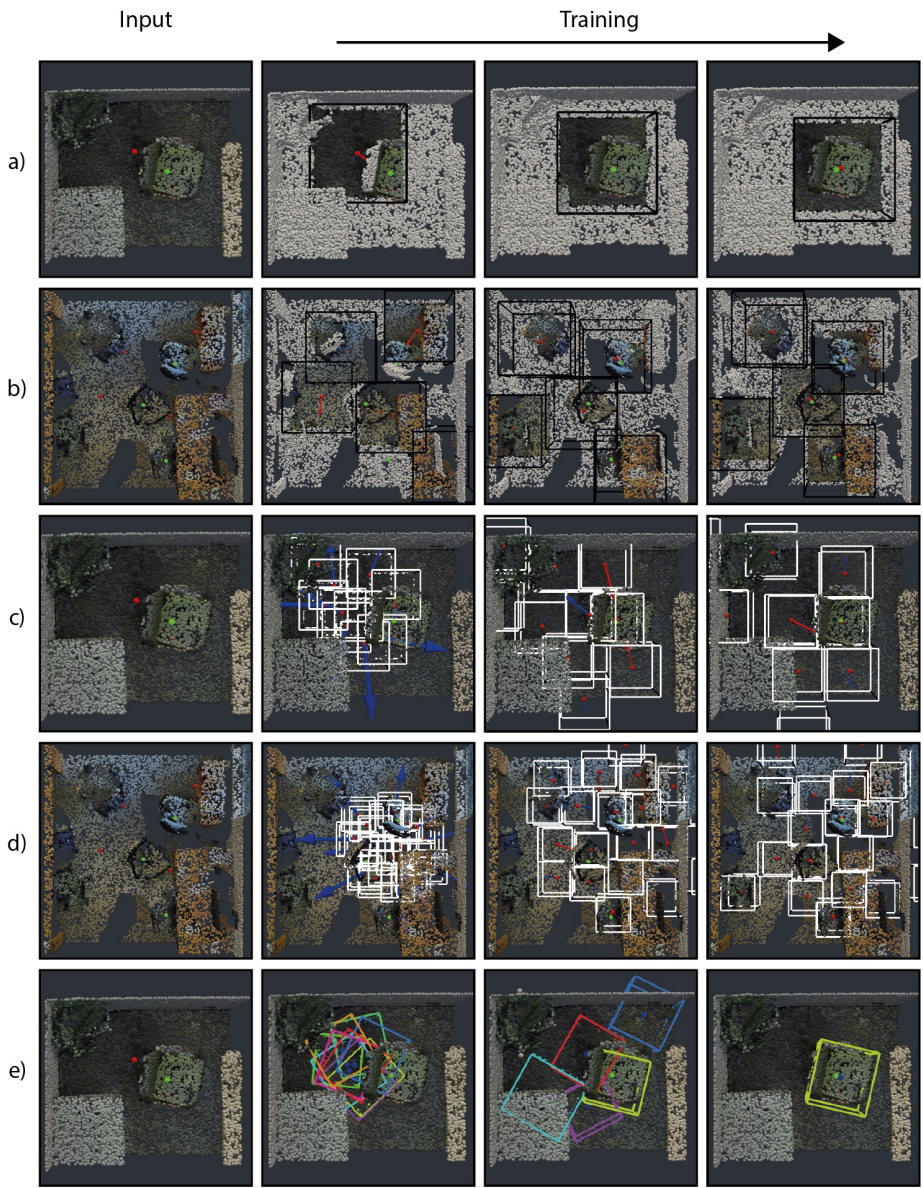
\myfigure{training_figure}{Visualizations for the \emph{scene network} during training. \textbf{a)} A scene consisting of a single proposal and single object. Colored points indicate those passed seen by the \emph{loss network} for gradient estimation (red arrow). \textbf{b)} A scene consisting of 5 proposals and 5 objects. \textbf{c)} Here we show 10 proposals with 1 object. Our overlap penalty (blue arrow) is applied to all but the closest proposal to an object. \textbf{d)} 20 proposals with 5 objects. \textbf{e)} 10 proposals with 1 object. Here we visualize the final scene network output which consists of location, bounding box and orientation. Proposals with an objectness score below a threshold are shown as grey dots. For full video sequences visit the project page at: \url{dgriffiths3.github.io}.}

\end{document}